\documentclass[11pt]{article}

\usepackage[final]{acl}
\usepackage{times}
\usepackage{latexsym}

\usepackage[T1]{fontenc}

\usepackage[utf8]{inputenc}
\usepackage{microtype}
\usepackage{inconsolata}

\usepackage{graphicx}

\usepackage[whole]{bxcjkjatype}
\usepackage{booktabs}

\usepackage{algorithm}
\usepackage{algpseudocode}
\usepackage{multirow}
\usepackage{amsmath}
\usepackage{tabularx}
\usepackage{makecell}

\newcommand{\nbf}[1]{{\noindent\textbf{#1}.~}}

\title{
    \textbf{A11y-Compressor: A Framework for\\
Enhancing the Efficiency of GUI Agent Observations through\\
 Visual Context Reconstruction and Redundancy Reduction}
}

\author{
  \textbf{Michito Takeshita \quad Takuro Kawada \quad Takumi Ohashi} \\
  \textbf{Shunsuke Kitada \quad Hitoshi Iyatomi} \\
  Hosei University, Tokyo, Japan \\
  \small{
   \textbf{Correspondence:} 
   \href{mailto:michitotakeshita00@gmail.com}
   {\texttt{michitotakeshita00@gmail.com}}, 
   \href{mailto:iyatomi@hosei.ac.jp}
   {\texttt{iyatomi@hosei.ac.jp}}
  }
}

\begin{document}
\maketitle

\begin{abstract}

AI agents that interact with graphical user interfaces (GUIs) require effective observation representations for reliable grounding.
The accessibility tree is a commonly used text-based format that encodes UI element attributes, but it suffers from redundancy and lacks structural information such as spatial relationships among elements.
We propose A11y-Compressor, a framework that transforms linearized accessibility trees into compact and structured representations.
Our implementation, Compressed-a11y, applies a lightweight and structured transformation pipeline with modal detection, redundancy reduction, and semantic structuring.
Experiments on the OSWorld benchmark show that Compressed-a11y reduces input tokens to 22\% of the original while improving task success rates by 5.1 percentage points on average.

\end{abstract}

\section{Introduction}
\label{sec:introduction}

AI agents that interact with graphical user interfaces (GUIs) have advanced rapidly with multimodal large language models (MLLMs)~\cite{lin2025showui, hong2024cogagent}. These agents perform tasks by interpreting complex on-screen environments, such as booking flights or responding to emails. While cloud-hosted closed-source MLLMs achieve strong performance, their real-world deployment is constrained by privacy risks, latency, and operational costs~\cite{zhang2023appagent, zhang2024ufo}. As a result, locally deployed open-source MLLMs have emerged as a practical alternative~\cite{niu2024screenagent, wang2024mobileagentv2}. However, these models face significant challenges in grounding, i.e., aligning UI elements with executable actions~\cite{wu2025guiactor}.

\begin{figure}[t]
    \centering
    \includegraphics[trim={0.7cm 0 1cm 0},clip, width=\linewidth]{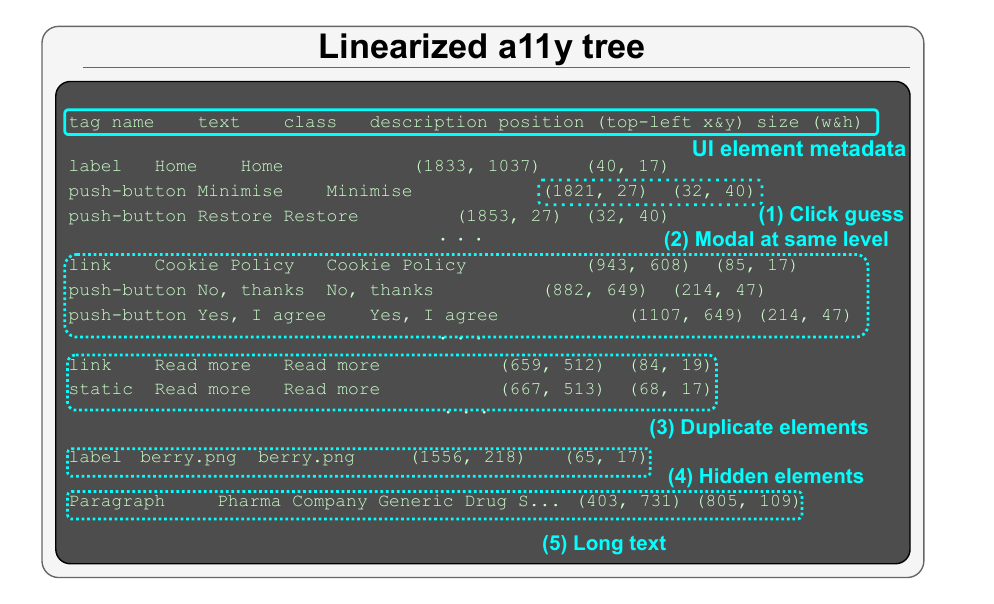}
    
    \caption{
        Examples of a linearized a11y tree, illustrating several issues:
        (1) providing position and size information can lead to unnecessary click-coordinate inference;
        (2) modal UIs are represented at the same hierarchical level as background elements, obscuring visual layering;
        (3) the same visual element may appear under multiple tags, creating ambiguity in element selection;
        (4) visually hidden elements are still included in the UI tree;
        and (5) elements with the Paragraph role tend to contain long text spans, increasing context length usage.
    }

    \label{fig:linearized_a11y}
\end{figure}

Effective grounding critically depends on how GUI environments are represented as observations~\cite{cheng2024seeclick, zheng2024seeact}. Existing approaches can be broadly categorized into image-based and text-based representations~\cite{xie2024osworld, zhou2023webarena}. Image-based representations provide rich visual information but often struggle with precise element localization~\cite{cheng2024seeclick, lin2025showui}. In contrast, text-based representations explicitly encode semantic attributes such as element roles, names, and positions, facilitating more reliable target identification~\cite{zhou2023webarena, xie2024osworld}.

The accessibility (a11y) tree is a used text-based representation that organizes UI elements hierarchically~\cite{zhou2023webarena, xie2024osworld}. Despite its effectiveness, it has two key limitations. First, its hierarchical structure does not align with the visual layout, making it difficult to capture spatial relationships and semantic regions~\cite{kerboua2025lineretriever}. Second, it contains substantial redundancy due to exhaustive attribute preservation, which increases token consumption and diffuses model attention~\cite{kerboua2025lineretriever, xie2024osworld}. Prior approaches, such as element selection and linearization, partially address redundancy but still fail to preserve spatial ordering and intrinsic GUI structures~\cite{kerboua2025lineretriever, deng2023mind2web, xie2024osworld}.
In particular, the linearized a11y tree is widely adopted as an observation representation for agents; however, it exhibits several challenges, as illustrated in Figure~\ref{fig:linearized_a11y}.

To address these limitations, we propose A11y-Compressor, a framework for constructing compact and structured GUI observations from linearized a11y trees. The framework consists of three stages: modal detection, which reconstructs foreground-background relationships; redundancy reduction, which removes irrelevant or repetitive elements; and semantic structuring, which organizes elements into meaningful groups. This design preserves essential structural information while significantly reducing token overhead, enabling more effective grounding for local MLLMs.

\nbf{Our main contributions}
 (1) We propose A11y-Compressor, a structured framework for constructing efficient GUI observation representations from the linearized a11y tree.
 (2) We demonstrate that observations generated by our framework significantly improve task success rates while reducing input token consumption for local MLLM-based GUI agents compared with existing observation formats on a GUI agent benchmark.

\begin{figure*}[t]
    \centering
    \includegraphics[trim={2cm 0 0.5cm 0},clip, width=\textwidth]{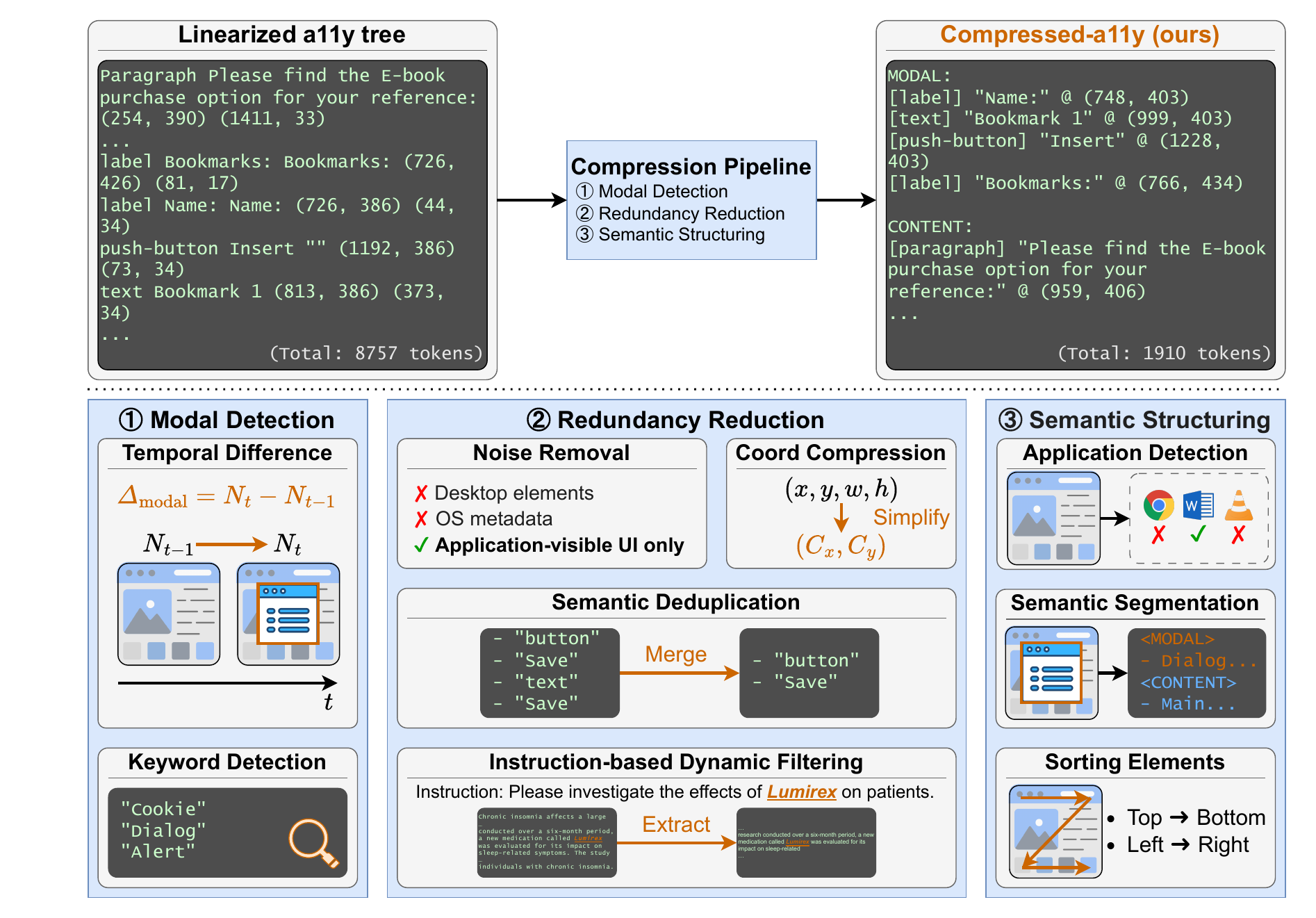}
    \caption{
        Overview of the A11y-Compresser framework. Given a linearized a11y tree, the pipeline applies modal detection, redundancy reduction, and semantic structuring to generate a compact and structure-preserving observation representation. 
        This representation enables more efficient and effective grounding for GUI agents.
    }
    \label{fig:methods}
\end{figure*}

\section{Related Work}
\label{sec:related_work}

\nbf{Observation Representations}
A central challenge in GUI agents is how to represent the environment for decision-making. Existing approaches can be broadly categorized into image-based and text-based representations. Image-based methods directly process screenshots using vision-language models, capturing rich visual context but often struggling with precise element localization~\cite{cheng2024seeclick, lin2025showui}. In contrast, text-based representations encode structured information such as element roles, attributes, and positions, enabling more accurate grounding~\cite{zhou2023webarena, xie2024osworld}. Hybrid approaches combining both modalities have also been explored, but they often incur higher computational costs.

\nbf{Accessibility Trees and Compression}
The accessibility (a11y) tree is a widely adopted text-based representation for GUI agents, as it provides a hierarchical view of UI elements and their attributes~\cite{zhou2023webarena, xie2024osworld}. Prior work has proposed various techniques to improve its efficiency, including element filtering, attribute selection, and linearization~\cite{kerboua2025lineretriever, deng2023mind2web, xie2024osworld}. These methods primarily focus on reducing redundancy and token usage. However, they often overlook the structural mismatch between hierarchical representations and the visual layout, leading to loss of spatial and semantic relationships. In contrast, our work focuses on constructing compact representations that preserve both structural and semantic information, enabling more effective grounding.
\section{A11y-Compressor}
\label{sec:methods}

As illustrated in Figure~\ref{fig:methods}, A11y-Compressor is a structured framework that transforms linearized a11y trees into compact, semantically coherent GUI observations.
Given a linearized a11y tree as input, the transformation is implemented as a three-phase pipeline: Modal Detection, Redundancy Reduction, and Semantic Structuring, each addressing a distinct aspect of observation construction.
The framework is modular, allowing each phase to be instantiated with different algorithms as long as they satisfy structural objectives.
The resulting representation preserves essential GUI structure while significantly reducing token overhead, enabling more efficient and reliable grounding.

%%%%%%%%%%%%%%%%%%%%%%%%%%%%%%%%%%%%%%%%%%%%%%% 

\subsection{Modal Detection}
\label{sec:methods:modal_detection}

This phase identifies foreground modal UI elements and separates them from background elements to make interaction constraints explicit.
Foreground elements such as modal UI elements (e.g., dialogs or pop-ups) introduce front–back relationships that restrict interaction with background elements; however, in the a11y tree, these elements are listed in parallel, potentially leading agents to select non-interactable background elements.

To address this issue, this phase detects modal UI elements and separates them from background elements, thereby constructing interaction constraints that arise from visual stacking relationships.
In practice, modal elements are identified based on accessibility attributes in the linearized a11y tree, such as tags and task-specific keywords (e.g., \textit{cookie}, \textit{accept}, or action-related labels). In addition, for modals triggered by interactions, newly appearing UI elements within the same screen state are identified as modal regions.
Elements that satisfy modal characteristics are assigned to the modal set $M$, while remaining elements are treated as background elements $B$.
Formally,
\begin{equation}
    (M, B) = f_{\text{modal}}(x),
\end{equation}
where $x$ denotes the linearized a11y tree, and $(M, B)$ are passed to the subsequent phase.

%%%%%%%%%%%%%%%%%%%%%%%%%%%%%%%%%%%%%%%%%%%%%%% 

\subsection{Redundancy Reduction}
\label{sec:methods:redundancy_reduction}

This phase reduces redundant and irrelevant information in the observation representation to improve the efficiency and reliability of grounding for GUI agents.
Linearized a11y trees often contain forms of redundancy, including duplicated UI elements, verbose textual content, and elements irrelevant to the current task (e.g., background or off-window components). 
These factors increase input length and hinder the agent's ability to identify correct interaction targets.
In addition, the spatial representation of UI elements can introduce ambiguity.
The linearized a11y tree represents each element using its top-left coordinate and bounding box size; however, the top-left coordinate does not always correspond to an effective interaction point.
To simplify spatial reasoning, this phase converts bounding box representations into center coordinates.

To address these issues, this phase applies rule-based preprocessing, including filtering irrelevant elements, merging duplicates, normalizing attributes, and compressing text.
Formally, given modal and background elements $(M,B)$ from the previous phase, it produces refined element sets:
\begin{equation}
    (M', B') = f_{\text{reduce}}(M, B),
\end{equation}
where $M'$ and $B'$ denote the modal and background elements after redundancy reduction. 
The refined element sets $M'$ and $B'$ are then passed to Section \ref{sec:methods:semantic_structuring} for semantic structuring.

%%%%%%%%%%%%%%%%%%%%%%%%%%%%%%%%%%%%%%%%%%%%%%% 

\subsection{Semantic Structuring}
\label{sec:methods:semantic_structuring}

This phase organizes UI elements into semantically meaningful regions to better reflect the functional structure of the GUI interface.
The linearized a11y tree often does not explicitly represent high-level semantic information about the GUI interface, such as which application it belongs to or what functional role each UI element plays.
As a result, GUI agents must implicitly infer the current screen context and the functional meaning of UI elements (e.g., which button performs which operation).

To address this, this phase augments the observation with explicit semantic structure.
This phase first identifies the application associated with the current interface, then partitions UI elements into semantic regions based on spatial layout and application-specific heuristics.
These regions correspond to coherent functional areas (e.g., taskbars or navigation panels).
Semantic regions are determined using application-specific heuristics derived from spatial layout and grounding patterns.
Formally, given the background elements $B'$ obtained from the previous phase, this phase reorganizes them into a set of semantic regions:
\begin{equation}
    R = f_{\text{region}}(B'),
\end{equation}
where $R = {r1, r2, …, rk}$ denotes the set of detected semantic regions.
The detected modal elements $M′$ are incorporated into the structured representation to construct the final observation used by the GUI agent:
\begin{equation}
    O = f_{\text{struct}}(R, M'),
\end{equation}
where $O$ denotes the semantically structured observation used for grounding and action selection.
\section{Experiments}
\label{sec:experiments}

%%%%%%%%%%%%%%%%%%%%%%%%%%%%%%%%%%%%%%%%%%%%%%% 

\subsection{Experiments Setup}
\label{ex:Experimental_Setup}
We evaluate the effectiveness of A11y-compressor on the GUI agent benchmark OSWorld~\cite{xie2024osworld}.
Our evaluation set consists of 358 tasks from the standard task set, excluding tasks that could not be executed due to environment-dependent errors.
The task distribution across application domains is as follows: web browsing (Chrome: 44), office work (LibreOffice Calc: 46, Impress: 47, Writer: 23), email management (Thunderbird: 15), media editing (GIMP: 26, VLC: 17), software development (VS Code: 23), basic OS operations (24), and cross-application tasks involving multiple applications (93).
This task set broadly covers practical GUI operation scenarios.
In all evaluation experiments, we employ Qwen3-VL-32B~\cite{bai2025qwen3vl} as the MLLM for inference.
Although we focus on Qwen3-VL-32B for controlled evaluation, the proposed representation is model-agnostic and applicable to other MLLMs that accept textual GUI observations.

%%%%%%%%%%%%%%%%%%%%%%%%%%%%%%%%%%%%%%%%%%%%%%% 

\subsection{Implementation of A11y-compressor}
\label{ex:Implementation_of_A11y-Compresser}

To evaluate the proposed framework, we implement A11y-Compressor using a rule-based approach.
Each phase is instantiated with heuristic rules derived from the structural characteristics of GUI interfaces, capturing common GUI patterns while remaining lightweight for efficient preprocessing.

\nbf{Modal Detection}
The modal detection phase identifies foreground UI elements that block interactions with background elements.
Modal elements are detected using two complementary strategies: temporal-difference detection and keyword-based detection.
Temporal-difference detection compares the linearized a11y tree at step $t$ with that at step $t-1$.
If the screen state remains unchanged but new UI elements appear, they are treated as modal candidates.
Keyword-based detection identifies UI elements containing representative modal-related keywords (e.g., cookie).
By combining these signals, foreground modal elements are separated from background elements.
Detailed rules are provided in Appendix~\ref{sec:appendix_modal_detection} and Appendix~\ref{app:keyword_modal}.

\nbf{Redundancy Reduction}
The redundancy reduction phase removes duplicated or irrelevant UI elements and compresses textual content in the observation representation.
It also converts bounding box representations into center coordinates to simplify spatial reasoning.
For textual compression, keywords are first extracted from the task instruction.
If a \texttt{paragraph} tag contains a matching keyword, the surrounding context is preserved; otherwise, only a predefined number of leading characters is retained.
Detailed implementation rules are provided in Appendix~\ref{appendix:redundancy_reduction}.

\nbf{Semantic Structuring}
The semantic structuring phase organizes UI elements into semantically coherent regions.
UI elements are first sorted from top-left to bottom-right based on their center coordinates, then partitioned into functional regions (e.g., APP\_LAUNCHER, CONTENT) using application-specific heuristics derived from spatial layout.
Detailed rules are provided in Appendix~\ref{app:semantic_structuring}.

To support the OSWorld benchmark~\cite{xie2024osworld}, we instantiate each phase using rule-based heuristics designed from 145 screen states across nine application domains (e.g., Chrome, Writer, VS Code).
These heuristics rely solely on structural and visual characteristics, without task-specific tuning, and are derived from data independent of the evaluation set to avoid benchmark bias.

%%%%%%%%%%%%%%%%%%%%%%%%%%%%%%%%%%%%%%%%%%%%%%% 

\subsection{Baseline Methods}
\label{ex:Baseline_methods}

We evaluate Compressed-a11y, an observation representation generated by A11y-Compressor, and compare it with three commonly used baselines.
\textbf{(1) Screenshot.}
A raw GUI screenshot directly provided as input to the MLLM.
\textbf{(2) Linearized a11y tree.}
A textual representation obtained by linearizing the hierarchical a11y tree into a one-dimensional sequence.
\textbf{(3) LineRetriever~\cite{kerboua2025lineretriever}.}
A method that dynamically selects task-relevant lines from the a11y tree based on their contribution to actions.
We extend the original web-based method to multiple application domains and use a lightweight, low-latency LLM (Qwen3-4B~\cite{yang2025qwen3}) as the retriever.

For quantitative evaluation, we compare Compressed-a11y with Linearized a11y Tree and LineRetriever in terms of success rate and token efficiency.
Screenshot-based observations are evaluated only in terms of success rate due to their distinct token characteristics.
For qualitative analysis of modal interaction, we compare Compressed-a11y with Screenshot and Linearized a11y Tree to examine their impact on agent reasoning and actions.
Additional qualitative results for LineRetriever are provided in Appendix~\ref{app:Case_study}.

%%%%%%%%%%%%%%%%%%%%%%%%%%%%%%%%%%%%%%%%%%%%%%% 

\subsection{Evaluation Metrics}
\label{ex:Evaluation_Metrics}

We evaluate agent performance using two quantitative metrics and a qualitative case study.
\textbf{Success Rate (SR).}
SR is defined as the proportion of tasks successfully completed within a maximum of 15 interaction steps.
To mitigate the impact of non-deterministic factors in the OSWorld environment, such as system response delays and variability in MLLM outputs, we conduct two trials for each task and consider a task successful if at least one trial succeeds.
\textbf{Average Input Tokens.}
We measure the average number of input tokens provided to the MLLM per application domain to quantify the efficiency of the observation representation.
\textbf{Case Study.}
In addition to quantitative evaluation, we analyze the behavior of different observation representations through a qualitative case study involving a task with a modal dialog.

%%%%%%%%%%%%%%%%%%%%%%%%%%%%%%%%%%%%%%%%%%%%%%% 

\subsection{Ablation Study}
\label{ex:Ablation_Study}
To analyze the contribution of each phase of the A11y-compressor framework, we conduct an ablation study.
We compare the full three-phase pipeline with variants that apply each phase individually, including modal detection, redundancy reduction, and semantic structuring.

\section{Results}
\label{sec:results}

%%%%%%%%%%%%%%%%%%%%%%%%%%%%%%%%%%%%%%%%%%%%%%% 

\subsection{Quantitative Results}
\label{res:Quantitative_Results}
\begin{figure*}[!t]
    \centering
    \includegraphics[width=\textwidth]{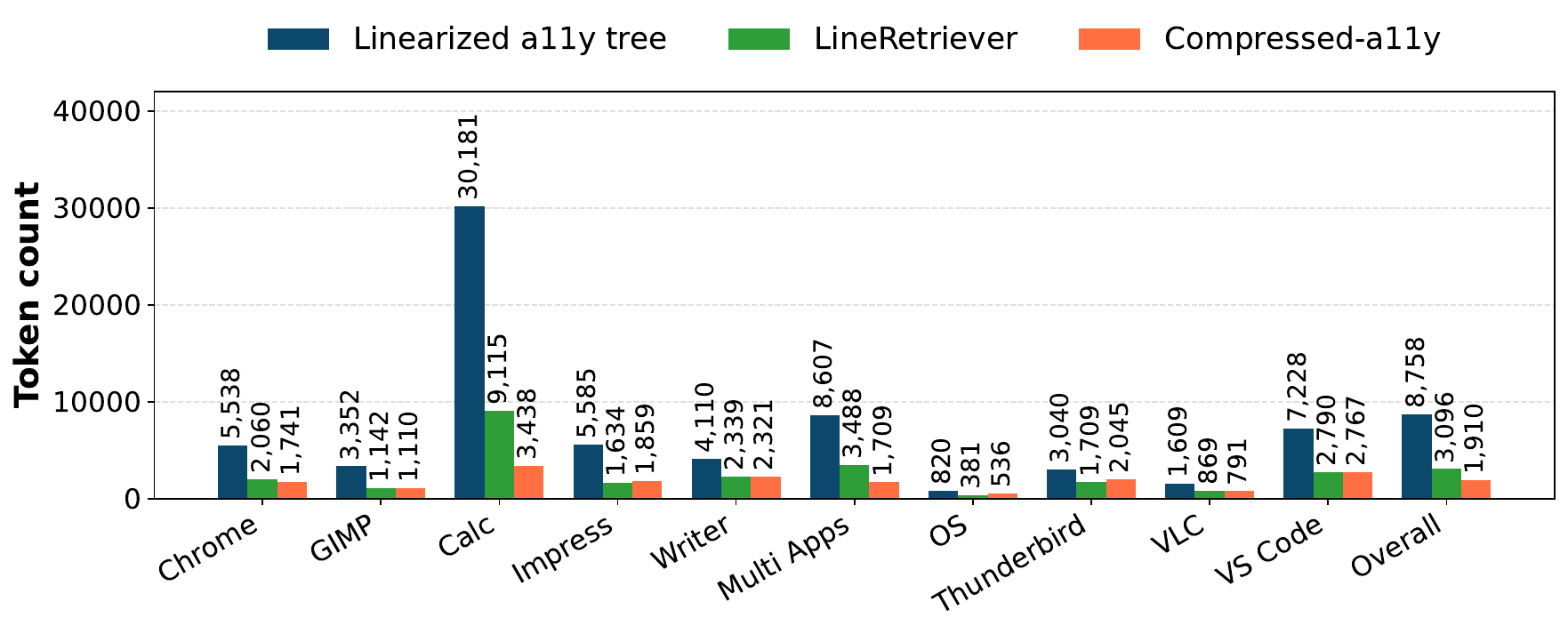}
    \caption{
        Average input token counts of observation representations across application domains.
        Compressed-a11y consistently achieves lower input token counts than both the linearized a11y tree and LineRetriever across domains, while maintaining a compact and stable representation even for applications with complex user interfaces.
    }
    \label{fig:tokens}
\end{figure*}

\nbf{Token Efficiency}
Figure~\ref{fig:tokens} compares the average number of input tokens for each observation representation across application domains.
Overall, while LineRetriever effectively reduces the average input token count compared to the linearized a11y tree, it still produces large input sizes in domains with inherently high token complexity.
In contrast, Compressed-a11y consistently limits the number of input tokens to approximately 3{,}500 or fewer across all application domains.
These results demonstrate that Compressed-a11y effectively suppresses token growth even for large-scale and complex UIs, enabling more efficient processing by local MLLMs.

\nbf{Task Success Rate}
Table~\ref{tab:success_rate} reports task success rates across the different observation representations.
Compressed-a11y achieves the highest overall average success rate (0.207), outperforming all baseline representations.
In particular, it achieves success rates of 0.304 and 0.467 for LibreOffice Writer and Thunderbird, respectively, showing substantial improvements over the baselines in these domains.
Although LineRetriever improves performance over the linearized a11y tree in several domains, its average success rate remains slightly lower than the linearized a11y tree baseline.

\begin{table*}[t]
    \centering
    \resizebox{\textwidth}{!}{%
    \begin{tabular}{lccccccccccc}
        \toprule
        \textbf{Method} & \textbf{Chrome} & \textbf{GIMP} & \textbf{Calc} & \textbf{Impress} & \textbf{Writer} & \textbf{Multi Apps} & \textbf{OS} & \textbf{Thunderbird} & \textbf{VLC} & \textbf{VS Code} & \textbf{Overall} \\
        \midrule
        Screenshot & 0.045 & 0.115 & 0.000 & 0.021 & 0.043 & \textbf{0.108} & 0.208 & 0.000 & 0.118 & 0.043 & 0.070 \\
        Linearized a11y tree & 0.182 & 0.192 & 0.000 & 0.149 & 0.087 & \textbf{0.108} & 0.333 & 0.267 & \textbf{0.294} & 0.304 & 0.156 \\
        LineRetriever & 0.136 & 0.192 & 0.022 & \textbf{0.191} & 0.087 & \textbf{0.108} & 0.333 & 0.133 & 0.176 & \textbf{0.348} & 0.151 \\
        \textbf{Compressed-a11y (ours)} & \textbf{0.250} & \textbf{0.231} & \textbf{0.043} & \textbf{0.191} & \textbf{0.304} & \textbf{0.108} & \textbf{0.375} & \textbf{0.467} & \textbf{0.294} & \textbf{0.348} & \textbf{0.207} \\
        \bottomrule
    \end{tabular}%
    }
    \caption{
        Task success rates across various application domains. Compressed-a11y consistently achieves the highest or tied-highest success rates across most domains. The highest score for each domain is highlighted in bold.
    }
    \label{tab:success_rate}
\end{table*}

%%%%%%%%%%%%%%%%%%%%%%%%%%%%%%%%%%%%%%%%%%%%%%% 

\subsection{Case Study: Modal Dialog Handling}
\label{res:Case_Study}
Figure~\ref{fig:case_study} presents a representative case study of a task involving a modal dialog.
The agent must correctly handle a privacy consent modal before interacting with the underlying flight booking interface.
With screenshot-based observations, the agent recognizes the modal but generates inaccurate click coordinates, leading to interactions at incorrect locations.
As a result, it repeatedly fails to interact with the intended UI elements and eventually reaches the maximum step limit.
With the linearized a11y tree, the agent fails to recognize that the modal blocks interaction with the background interface and attempts to interact with elements behind it, again resulting in failure.

In contrast, Compressed-a11y successfully identifies the modal and its relevant UI elements.
After initially attempting to close the dialog, the agent recognizes that additional actions are required and enables the necessary toggles before confirming the dialog.
This structured representation allows the agent to correctly complete the modal interaction and proceed with the task.

\begin{figure*}[t]
    \centering
    \includegraphics[trim={0.8cm 0 0 0},clip, width=\textwidth]{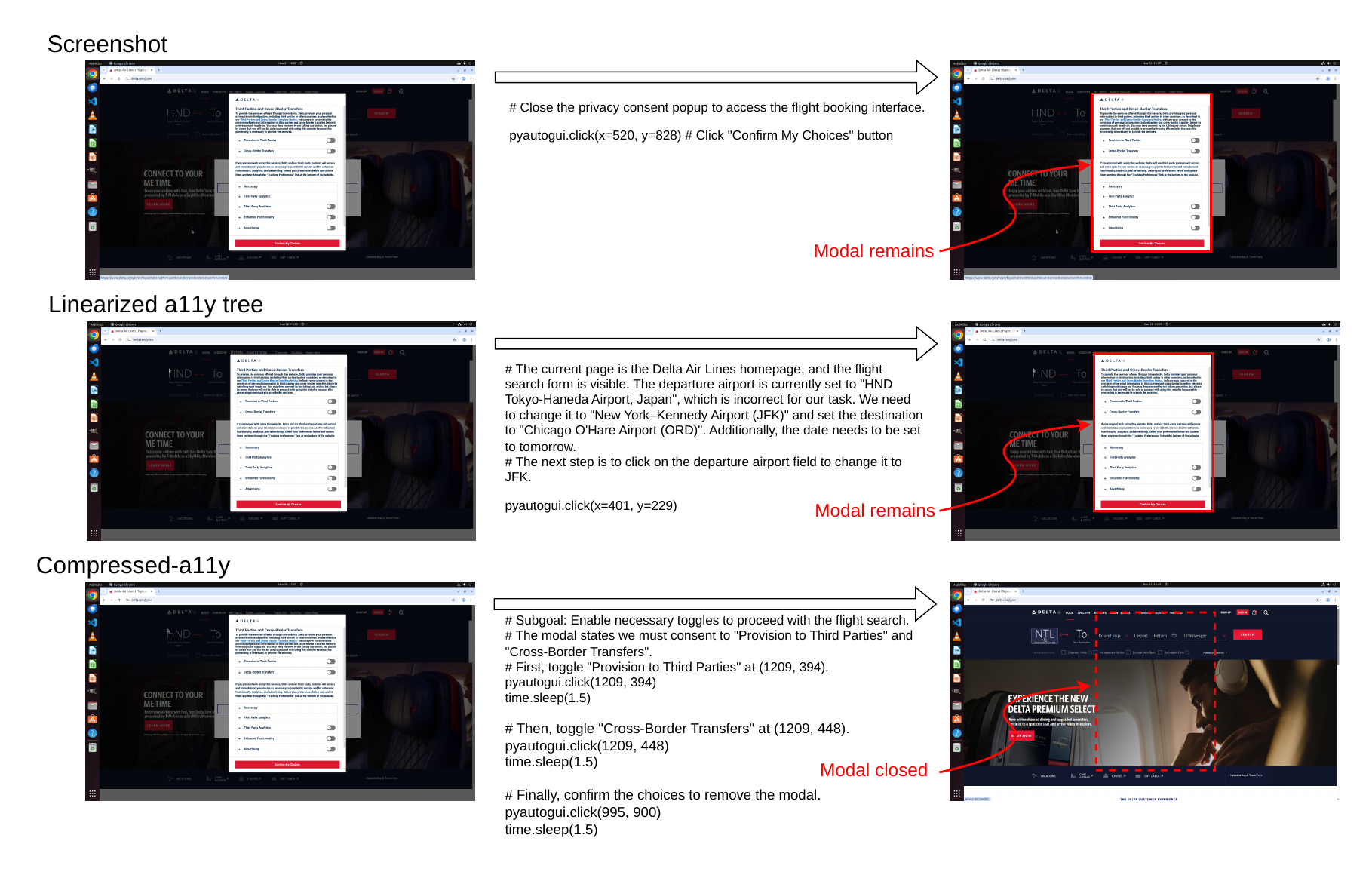}
    \caption{
        Example of modal dialog handling with different observation representations.
        Screenshot-based observations suffer from inaccurate coordinate grounding leading to incorrect interactions.
        The linearized a11y tree fails to capture modal-induced interaction constraints and attempts to interact with background elements.
        In contrast, Compressed-a11y explicitly models the modal structure, enabling correct interaction sequencing and successfully successful task completion.
    }
    \label{fig:case_study}
\end{figure*}

%%%%%%%%%%%%%%%%%%%%%%%%%%%%%%%%%%%%%%%%%%%%%%% 

\subsection{Ablation Study}
\label{res:Case_Study}
Table~\ref{tab:ablation_study} presents a component-wise analysis of the A11y-Compresser framework.
The resulting representation (Compressed-a11y) achieves the highest overall success rate (0.207) and outperforms all ablated variants acrossamong mostall configurations.application domains.
The redundancy reduction–only variant achieves performance comparable to the linearized a11y tree baseline (0.156).
In particular, it attains a success rate of 0.467 on Thunderbird, matching the full pipeline in this domain.
Both modal detection only and semantic structuring only variants achieve lower overall success rate (0.134).
While modal detection shows relatively stable performance across  domains, semantic structuring exhibits more domain-dependent behavior.
For example, semantic structuring achieves the highest success rate in Calc and Multi Apps, but fails to solve any tasks in the Thunderbird domain.
These results show that combining all three phases leads to the most consistent performance across application domains.

\begin{table*}[t]
    \centering
    \resizebox{\textwidth}{!}{%
    \begin{tabular}{lccccccccccc}
        \toprule
        \textbf{Method} & \textbf{Chrome} & \textbf{GIMP} & \textbf{Calc} & \textbf{Impress} & \textbf{Writer} & \textbf{Multi Apps} & \textbf{OS} & \textbf{Thunderbird} & \textbf{VLC} & \textbf{VS Code} & \textbf{Overall} \\
        \midrule
        Full pipeline (Compressed-a11y)  & 0.250 & 0.231 & 0.043 & 0.191 & 0.304 & 0.108 & 0.375 & 0.467 & 0.294 & 0.348 & 0.207 \\
        Modal detection only  & 0.159 & 0.077 & 0.022 & 0.191 & 0.087 & 0.108 & 0.250 & 0.200 & 0.118 & 0.261 & 0.134 \\
        Redundancy reduction only  
        & 0.182 & 0.154 & 0.043 & 0.191 & 0.130 & 0.108 & 0.167 & 0.467 & 0.176 & 0.261 & 0.156 \\
        Semantic structuring only
        & 0.159 & 0.115 & 0.109 & 0.170 & 0.000 & 0.118 & 0.333 & 0.000 & 0.059 & 0.217 & 0.134 \\
        \bottomrule
    \end{tabular}%
    }
    \caption{
        Ablation study of the proposed A11y-Compresser framework across application domains.
        We compare the full three-phase pipeline (Compressed-a11y) with variants that apply each phase individually, including modal detection, redundancy reduction, and semantic structuring.
        Each column reports the task success rate for the corresponding application domain, highlighting the best-performing configuration and enabling direct comparison of the contribution of each component.
    }
    \label{tab:ablation_study}
\end{table*}

\section{Discussion}
\label{sec:discussion}

%%%%%%%%%%%%%%%%%%%%%%%%%%%%%%%%%%%%%%%%%%%%%%%

\subsection{Effectiveness of Compressed-a11y}
\label{dis:Compressed-a11y}
Compressed-a11y consistently achieves the highest task success rate across most tasks while reducing the number of input tokens to approximately 22\% of that required by the baseline linearized a11y tree, corresponding to an average improvement of 5.1 percentage points.
Token reduction rates varied across application domains and were closely related to the complexity of the UI in each domain.

The linearized a11y tree contain large amounts of redundant accessibility information, particularly in application domains with complex UIs such as Calc and VS Code.
By aggregating and compressing such redundant information, Compressed-a11y achieves more pronounced token reductions in these domains.
Meanwhile, in application domains with smaller reduction rates, such as OS and VLC, the number of UI elements was inherently limited and the redundancy of accessibility information was relatively low; consequently, the impact of compression was correspondingly smaller.
Notably, despite variations in compression rate, Compressed-a11y maintains the highest task success rate across most application domains.

These results indicate that Compressed-a11y effectively extracts task-relevant information while preserving essential GUI structure, enabling more efficient screen understanding and interaction by MLLMs.
Furthermore, the effectiveness of Compressed-a11y varies across application domains.
This variation likely arises because the compression pipeline incorporates application-specific designs, such as UI region segmentation based on structural characteristics of each interface.
When these designs align well with the UI characteristics of a given application, Compressed-a11y yields larger performance gains; otherwise, the improvements are more limited.
This observation suggests that considering application-specific UI layout characteristics is important for effectively organizing and highlighting task-relevant elements in compressed observation representations.

%%%%%%%%%%%%%%%%%%%%%%%%%%%%%%%%%%%%%%%%%%%%%%%

\subsection{Modal Interaction Analysis}
\label{dis:Modal_interaction_analysis}

The qualitative case study highlights the importance of explicitly representing modal dialogs in observation representations.
In screenshot-based observations, the agent often fails due to imprecise coordinate grounding, while the linearized a11y tree does not explicitly distinguish modal elements from background UI components. 
As a result, the agent may attempt to interact with elements that are temporarily inaccessible, leading to failed interactions.
These observations indicate that observation representations should explicitly encode modal dialogs and provide clear grounding information for UI elements within the modal interface.
Without such modal-aware representations, agents may incorrectly attempt to interact with background elements that are temporarily inaccessible.

%%%%%%%%%%%%%%%%%%%%%%%%%%%%%%%%%%%%%%%%%%%%%%%

\subsection{Comparisons to Other Approaches}
\label{dis:comparisons_to_other_approaches}

Through comparative experiments, we observed that even with a strong Local-MLLM, relying solely on screenshots is insufficient for solving general GUI tasks. 
This finding underscores the necessity of structured observation representations.

The linearized a11y tree, which serves as the foundation of Compressed-a11y, achieved substantially higher task success rates than screenshots. 
However, as discussed earlier, for tasks involving complex UIs, the resulting observation representations tended to become redundant, potentially constraining the performance of the MLLM.

LineRetriever achieved a compact representation by using an LLM to extract only important lines. 
Nevertheless, it failed to preserve the global structural context required for GUI manipulation, resulting in inconsistent improvements in task success rates.
Because many tasks involve exploration, retrieval-based representations are prone to missing critical UI elements and their relationships in the early interaction stages. 
Furthermore, dependence on the inference results of the retriever LLM may have introduced variability in the extracted content, leading to unstable task performance.

By compressing observations while preserving task-relevant information, Compressed-a11y outperforms existing methods in both token efficiency and task success rate.
These results indicate that compressing the linearized a11y tree into a form that better captures overall UI structure improves GUI agent performance.

%%%%%%%%%%%%%%%%%%%%%%%%%%%%%%%%%%%%%%%%%%%%%%%

\subsection{Ablation Study }
\label{dis:Ablation_study}

The ablation analysis further clarifies the role of each phase in the A11y-Compressor framework. 
The full pipeline achieves the highest overall success rate, indicating that performance gains arise from the combination of multiple processing phases rather than any single component alone.

Among the single-phase variants, redundancy reduction only achieves the highest overall success rate.
A possible explanation is that converting UI coordinates to center-based positions reduces ambiguity in click actions.
When coordinates are represented by the top-left corner of a bounding box, clicking at that position may not always successfully interact with the intended UI element, depending on its visual shape.
In such cases, task success may depend on whether the model correctly infers a valid click position within the bounding box, introducing variability in task performance.
By converting coordinates to center-based positions, this ambiguity is reduced, leading to more reliable and consistent interactions.

In the Calc domain, semantic structuring only achieves the highest success rate. 
This may be because explicitly labeling interface regions helps the model focus on task-relevant UI components, reducing distraction from numerous irrelevant elements such as spreadsheet cells.

In contrast, although modal detection only successfully handles modal dialogs, the agent often fails in subsequent steps of the interaction. 
This limitation explains why modal detection alone does not substantially  improve overall task success.
\section{Conclusion}
\label{sec:conclusion}

In this paper, we propose A11y-Compressor, a framework for constructing compressed observation representations for GUI agents.
The framework transforms linearized a11y trees into a structured and compact representation, referred to as Compressed-a11y, by incorporating region segmentation and structural organization.
Experimental results on the OSWorld benchmark demonstrate that the proposed framework significantly reduces the number of input tokens to approximately 22\% of those required by the linearized a11y tree while improving task success rates across many application domains, achieving an average gain of 5.1 percentage points.
In particular, for applications with complex user interfaces, A11y-Compressor effectively suppresses redundant accessibility information while preserving the structural context necessary for successful task execution.
Future work will explore the applicability of the proposed representation to closed-MLLMs with larger capacity and stronger reasoning capabilities, as improvements with local MLLMs suggest similar gains.
\section{Limitations}
\label{sec:limitation}

Although A11y-Compressor is designed as a general framework for generating compressed GUI observation representations, several limitations remain in the current implementation.
The framework operates on the linearized a11y tree as its primary input representation. 
Consequently, A11y-Compressor cannot directly utilize visual information that is not represented in the accessibility tree, such as icon shapes, colors, or other purely visual cues. 
This limitation may reduce the effectiveness of the framework in tasks where such visual information plays a critical role in identifying UI elements.
In addition, our evaluation is conducted on representative desktop applications in the OSWorld benchmark. 
The applicability of the framework to other environments, such as mobile interfaces or different application ecosystems, has not yet been investigated.
Furthermore, the Compressed-a11y representation used in our empirical evaluation is implemented using rule-based procedures. 
As a result, several design choices, including threshold values, rely on heuristics, which may limit the robustness and generalizability of the current implementation across diverse interface settings.
Future work could extend A11y-Compressor by incorporating more flexible compression strategies, potentially including learning-based approaches, and by evaluating its effectiveness across a wider range of interface environments.

\bibliography{reference}

%%%%%%%%%%%%%%%%%%%%%%%%%%%%%%%%%%%%%%%%%%%%%%% 

\clearpage

\appendix
\section{Overview of Temporal Modal Detection}
\label{sec:appendix_modal_detection}
This section provides a detailed description of our implementation of the modal detection phase in A11y-Compressor.
In our implementation, we employ two complementary strategies: temporal-difference-based detection and keyword-based detection.
The temporal-difference-based method serves as the primary approach when consecutive observations correspond to the same screen.
In contrast, the keyword-based method is applied in situations where temporal correspondence is unavailable, such as the initial observation or during screen transitions.

We first describe the temporal-difference-based detection method.
We adopt a three-stage pipeline consisting of:
(1) temporal UI correspondence and same-screen identification,
(2) modal candidate extraction based on temporal differences, and
(3) rule-based modal validity scoring and decision.
Given two consecutive observations, we first determine whether they correspond to the same screen.
If so, newly appeared UI elements are treated as modal candidates and evaluated using a rule-based scoring scheme.

\subsection{Temporal UI Correspondence}
\label{app:temporal_correspondence}

Let $U^t=\{u_i^t \mid i = 1, 2, \dots, N_u^t\}$ denote the set of UI elements observed at time step $t$, where $N_u^t$ is the number of elements.
For each UI element $u_i^t$, we denote its position vector and semantic content (e.g., tag, name, text, class, and description) as $p_i^t$ and $c_i^t$, respectively.

To establish correspondence across consecutive observations, we define a semantic matching operator that pairs elements with identical semantic content:

{\small
\begin{multline}
\Gamma^t(U_A, U_B) =
\{(u_j^{t-1}, u_k^t)
\mid
u_j^{t-1}\in U_A,\;
u_k^t\in U_B, \\
c_j^{t-1}=c_k^t
\}.
\end{multline}
}

\subsection{Region-Aware Matching}
\label{app:region_matching}

UI elements exhibit different temporal behaviors depending on the region type.
We therefore partition UI elements into static and dynamic regions.
Let $U_\mathrm{STA}^t$ and $U_\mathrm{DYN}^t$ denote UI elements in static and dynamic regions, respectively.
Details of region detection are described in Section~\ref{sec:methods:semantic_structuring}.

\paragraph{Static Regions.}
UI elements in static regions are expected to remain spatially stable across consecutive observations.
We define the static matching indicator:

{\small
\begin{equation}
\mu_{\mathrm{STA}}^t(u_j^{t-1}, u_k^t) =
\begin{cases}
1, & \|p_j^{t-1} - p_k^t\| \le \epsilon_{\mathrm{STA}},\\
0, & \text{otherwise}.
\end{cases}
\end{equation}
}

Here, $\epsilon_{\mathrm{STA}}$ accounts for positional noise.

\paragraph{Dynamic Regions.}
In dynamic regions, UI elements may shift due to scrolling or viewport changes.
To compensate for this, we estimate a global displacement vector.

For each matched pair, we define:

{\small
\begin{equation}
\Delta p_{j,k}^t = p_k^t - p_j^{t-1}.
\end{equation}
}

The global translation is then estimated as:

{\small
\begin{equation}
\Delta p_\mathrm{GLB}^t =
\mathrm{median}
\big(
\{\Delta p_{j,k}^t\}
\big).
\end{equation}
}

Using this, the dynamic matching indicator is defined as:

{\small
\begin{equation}
\mu_{\mathrm{DYN}}^t(u_j^{t-1}, u_k^t) =
\begin{cases}
1, & \|p_j^{t-1} + \Delta p_{\mathrm{GLB}}^t - p_k^t\| \le \epsilon_{\mathrm{DYN}},\\
0, & \text{otherwise}.
\end{cases}
\end{equation}
}

\subsection{Same-Screen Identification}
\label{app:samescreen}

We determine whether two consecutive observations correspond to the same screen using a matching ratio over dynamic-region elements:

{\small
\begin{equation}
R^t =
\frac{1}{|U_\mathrm{DYN}^{t-1}|}
\sum_{(u_j^{t-1},u_k^t)\in
\Gamma^t(U_\mathrm{DYN}^{t-1}, U_\mathrm{DYN}^t)}
\mu_\mathrm{DYN}^t(u_j^{t-1},u_k^t).
\end{equation}
}

If $R^t$ exceeds a predefined threshold, the two observations are regarded as belonging to the same screen.

\paragraph{Rationale for the Denominator.}
We use $U_\mathrm{DYN}^{t-1}$ as the denominator to verify whether previously visible UI elements persist.
Using $U_\mathrm{DYN}^{t}$ would cause large modal overlays to artificially reduce the matching ratio.

\paragraph{Threshold Parameters.}
\begin{itemize}
    \item \textbf{Position tolerances ($\epsilon_{\mathrm{STA}}, \epsilon_{\mathrm{DYN}}$):} 25 pixels.
    \item \textbf{Matching ratio threshold:} 0.3.
\end{itemize}

\paragraph{Exception Handling.}
\begin{itemize}
    \item \textbf{Large modal handling:}
    If the number of matched elements exceeds 10, the screen is regarded as identical regardless of $R^t$.
    \item \textbf{Sparse screen handling:}
    If $|U^t| < 15$, we bypass same-screen judgment and proceed directly to modal detection.
\end{itemize}

\subsection{Temporal Difference-based Candidate Extraction}
\label{app:candidate_extraction}

If two observations are determined to belong to the same screen, modal candidates are extracted as newly appeared UI elements.

{\small
\begin{multline}
M^t =
\Big\{
u_i^t\in U^t \;\Big|\;
\forall u_j^{t-1}\in U_\mathrm{DYN}^{t-1},\;
\mu_{j,i}^{\mathrm{DYN},t}=0 \\
\land
\forall u_j^{t-1}\in U_\mathrm{STA}^{t-1},\;
\mu_{j,i}^{\mathrm{STA},t}=0
\Big\}.
\end{multline}
}

These elements represent UI components that newly emerge at time step $t$.

\subsection{Modal Validity Scoring}
\label{app:modal_score}

Let $M^t$ denote a modal candidate set and $m \in M^t$ an individual UI element.
We define the total modal score as:

{\small
\begin{equation}
M_{\mathrm{total}}
=
\sum_{m \in M^t}
\Bigl(
M_{\mathrm{tag}}(m)
+
M_{\mathrm{name}}(m)
\Bigr)
+
M_{\mathrm{count}}(M^t).
\end{equation}
}

\subsubsection{Tag-based Score}

{\small
\begin{equation}
M_{\mathrm{tag}}(m)=
\begin{cases}
+2.0 & \text{if } \mathrm{tag}(m) \in \mathcal{R}_{\mathrm{interactive}}, \\
-0.5  & \text{if } \mathrm{tag}(m) \in \mathcal{R}_{\mathrm{decorative}}, \\
0.0   & \text{otherwise}.
\end{cases}
\end{equation}
}

$\mathcal{R}_{\mathrm{interactive}}$ = \{dialog, alertdialog, menu, listbox, tree\}  
$\mathcal{R}_{\mathrm{decorative}}$ = \{image, label, heading, paragraph, generic\}

\subsubsection{Name-based Score}

{\small
\begin{equation}
M_{\mathrm{name}}(m)=
\begin{cases}
w_{\mathrm{decide}}, & \text{if }\;
\begin{array}[t]{@{}l@{}}
\mathrm{is\_interactive}(m)=1\\
\land\;\mathrm{name}(m)\cap \mathcal{K}_{\mathrm{decide}} \neq \emptyset
\end{array}
\\[2pt]
w_{\mathrm{func}}, & \text{if }\;
\begin{array}[t]{@{}l@{}}
\mathrm{is\_interactive}(m)=1\\
\land\;\mathrm{name}(m)\cap \mathcal{K}_{\mathrm{func}} \neq \emptyset
\end{array}
\\[2pt]
0.0, & \text{otherwise}.
\end{cases}
\label{eq:mtext}
\end{equation}
}

$\mathcal{K}_{\mathrm{decide}}$ = \{OK, Cancel, Save, Yes, No, Login, Agree, Delete\}  
$\mathcal{K}_{\mathrm{func}}$ = \{Sort, Filter, Settings, Search, Find\}

\subsubsection{Cardinality-based Correction}

{\small
\begin{equation}
M_{\mathrm{count}}(M^t)=
\begin{cases}
-3.0 & \begin{aligned}[t]
       &\text{if } |M^t| < 3 \\
       &\land \forall m \in M^t,\ M_{\mathrm{tag}}(m) \le 0
       \end{aligned} \\
+1.0 & \text{if } |M^t| \ge 6 \\
0.0  & \text{otherwise}.
\end{cases}
\end{equation}
}

\subsection{Final Decision Rule}
\label{app:decision_rule}

A modal candidate is accepted as a valid modal if:

{\small
\begin{equation}
\mathrm{is\_modal}(M^t)=
\begin{cases}
1 & \text{if } M_{\mathrm{total}} \ge T_{\mathrm{modal}}, \\
0 & \text{otherwise}.
\end{cases}
\end{equation}
}

In all experiments, we set $T_{\mathrm{modal}} = 1.0$.

%%%%%%%%%%%%%%%%%%%%%%%%%%%%%%%%%%%%%%%%%%%%%%% 

\section{Keyword-Based Modal Detection}
\label{app:keyword_modal}

We next describe the keyword-based detection method.
When temporal correspondence is unavailable, we apply this method as a complementary strategy to identify modal elements.
Typical examples include cookie consent banners, login dialogs, and informational pop-ups.

This method follows a three-stage pipeline:
(1) anchor extraction based on keywords,
(2) spatial clustering of anchor elements, and
(3) region-level scoring and decision.

\subsection{Search Region and Anchor Extraction}
\label{app:keyword_anchor}

Since modals appearing in initial states tend to emerge in characteristic regions
depending on the application domain, we first define a heuristic search region.
Let $O = \{ o_i \mid i = 1, 2, \dots, N_o \}$ denote the set of UI elements within this region.
For each element $o_i$, we denote its screen position vector as $p_i$.

We extract anchor elements based on predefined keyword sets.
An element is regarded as an anchor if its textual content contains any keyword in $K$.
The resulting anchor set is denoted as $A \subset O$.

\paragraph{Keyword Sets.}
We define two types of keywords:

\begin{itemize}
    \item \textbf{Content Keywords ($K_{\mathrm{content}}$):}
    \texttt{cookie}, \texttt{cookies}, \texttt{gdpr}, \texttt{privacy}, \texttt{consent}
    
    \item \textbf{Action Keywords ($K_{\mathrm{action}}$):}
    \texttt{accept}, \texttt{agree}, \texttt{allow}, \texttt{reject}, \texttt{save}, \texttt{confirm}, \texttt{close}, \texttt{×}, \texttt{ok}, \texttt{policy}, \texttt{manage}, \texttt{setting}
\end{itemize}

\subsection{Spatial Clustering of Anchor Elements}
\label{app:keyword_clustering}

To construct modal candidate regions, we group anchor elements based on spatial proximity.

Two anchor elements $a_j, a_k \in A$ are considered connected if:

{\small
\begin{equation}
\| p_j - p_k \| < \delta,
\end{equation}
}

where the distance threshold $\delta$ is defined as:

{\small
\begin{equation}
\delta = 0.08 \cdot \min(W, H),
\end{equation}
}

with $W$ and $H$ denoting the screen width and height.
Connected components formed under this criterion are treated as modal candidate regions.

\subsection{Edge-Based Immediate Detection}
\label{app:keyword_edge}

To efficiently detect cookie banners and similar notifications,
we introduce a geometric shortcut for edge-aligned regions.

A candidate region is immediately classified as a modal if it satisfies:

\begin{itemize}
    \item \textbf{Vertical Position:}
    \begin{equation}
        y > 0.75H \quad \lor \quad y < 0.15H
    \end{equation}

    \item \textbf{Aspect Ratio:}
    \begin{equation}
        \frac{w}{h} > 2.5
    \end{equation}
\end{itemize}

These conditions capture horizontally elongated regions located at the top or bottom of the screen,
which are typical for consent banners.

\subsection{Region-level Scoring}
\label{app:keyword_scoring}

For candidate regions not detected by the edge-based rule,
we evaluate modal validity using a composite score:

\begin{itemize}
    \item \textbf{Anchor Count Score:}
    Based on the number of anchor elements (capped at 20).
    
    \item \textbf{Centrality Score:}
    Based on distance from the screen center (maximum 30 points).
    
    \item \textbf{Structural Score:}
    Based on the presence of interactive UI components such as buttons, input fields, toggles, or close icons.
\end{itemize}

The total score is denoted as $S_{\mathrm{total}}$.

\subsection{Decision Rule}
\label{app:keyword_decision}

A candidate region is classified as a modal if:

{\small
\begin{equation}
S_{\mathrm{total}} \ge 65.0
\quad \land \quad
S_{\mathrm{total}} \ge 0.8 \times S_{\mathrm{max}},
\end{equation}
}

where $S_{\mathrm{max}}$ is the maximum score among candidates in the current frame.

\subsection{Rejection Criteria}
\label{app:keyword_rejection}

Even if a candidate satisfies the scoring conditions, it is rejected if:

\begin{itemize}
    \item The number of anchor elements is too small.
    \item The region area is too small.
    \item The region corresponds to a navigation bar or search form.
    \item The region lacks a clear closing or cancel mechanism.
    \item The region covers most of the screen, indicating a full page transition.
\end{itemize}

%%%%%%%%%%%%%%%%%%%%%%%%%%%%%%%%%%%%%%%%%%%%%%%

\section{Redundancy Reduction Process}
\label{appendix:redundancy_reduction}
This section provides a detailed description of our implementation of the redundancy reduction phase in A11y-Compressor.
We apply rule-based preprocessing steps, including UI element merging, attribute compression, and dynamic selection, to all UI elements in order to enhance the information density of the observation representation while improving MLLM inference efficiency.
Table~\ref{tab:preprocessing_steps} summarizes these preprocessing steps.
Further details on visual and semantic deduplication, as well as instruction-aware filtering, are provided in~\ref{app:deduplication} and~\ref{app:paragraph_compression}.

\begin{table*}[t]
\centering
\small
\begin{tabularx}{\textwidth}{@{} p{3.5cm} X X @{}}
\toprule
\textbf{Processing Step} & \textbf{Motivation} & \textbf{Processing Overview} \\
\midrule
\textbf{Rule-based Noise Removal} & 
To eliminate non-essential hidden elements and OS metadata that bloat the MLLM input context without contributing to task completion. & 
Filters out background desktop elements and system-level metadata using heuristic rules. \\ 
\addlinespace

\textbf{Visual \& Semantic}\\ \textbf{Deduplication} & 
To resolve redundancy in the linearized a11y tree where a single UI element is represented by multiple tags, reducing inference load. & 
Merges spatially overlapping or identical elements, prioritizing interactive tags (e.g., buttons) over static containers. \\ 
\addlinespace

\textbf{Attribute Selection \&}\\ \textbf{Coordinate Compression} &
To simplify decision-making by removing excessive geometric details that induce unnecessary reasoning. & 
Converts bounding boxes $(x,y,w,h)$ into center coordinates $(c_x, c_y)$ and retains only essential attributes ($\text{tag}, \text{name}$). \\ 
\addlinespace

\textbf{String Normalization} & 
To prevent matching failures caused by inconsistent formatting, whitespace, or unnecessary line breaks. & 
Normalizes whitespace and removes redundant newlines in attribute values to ensure consistency. \\ 
\addlinespace

\textbf{Instruction-based}\\ \textbf{Dynamic Filtering} & 
To avoid distracting the agent with long, irrelevant text paragraphs unrelated to the current task. & 
Dynamically truncates paragraph text based on keywords extracted from user instructions, retaining only relevant segments. \\
\bottomrule
\end{tabularx}
\caption{Overview of preprocessing steps for observation space reduction.}
\label{tab:preprocessing_steps}
\end{table*}

\subsection{UI Element Deduplication}
\label{app:deduplication}
To reduce the inference load on the MLLMs, we implement a preprocessing step that merges spatially and semantically overlapping UI elements.
This process eliminates redundant information while preserving interactive components.
The specific logic is implemented as follows.

\nbf{Deduplication Criteria}
Two UI elements are considered "duplicate candidates" if they satisfy both of the following conditions:

\begin{itemize}
    \item \textbf{Spatial Proximity:}
    We calculate the center coordinates $(c_x, c_y)$ of the bounding box for each element and measure the Euclidean distance between them.
    If this distance is within a predefined threshold (default: 20.0 pixels), the elements are judged to be in close proximity.
    \textit{Exception:} If the \texttt{name} attributes match exactly, we relax the condition to allow a vertical deviation ($y$-axis) of up to 30 pixels, tolerating larger horizontal deviations.

    \item \textbf{Semantic Similarity:}
    We compare the \texttt{name} attributes after normalization (lowercase conversion and whitespace removal).
    Elements are considered semantically similar if there is an exact match or if one string is a substring of the other.
    \textit{Prevention of Over-merging:} To prevent incorrect merging, if the lengths of the labels differ significantly (e.g., one is more than twice the length of the other), we regard them as distinct meanings and skip the merger.
\end{itemize}

\nbf{Tag Priority Strategy}
When two elements are identified as duplicate candidates, we prioritize retaining the more interactive element.
We define a priority score based on the element's \texttt{tag} (lower values indicate higher priority), as shown in Table~\ref{tab:tag_priority}.

\begin{table*}[t]
\centering
\small
\begin{tabularx}{\textwidth}{@{} c c X X @{}}
\hline
\textbf{Priority} & \textbf{Score} & \textbf{Tags} & \textbf{Rationale} \\
\hline
Highest & 0 &
\texttt{entry}, \texttt{combo-box}, \texttt{check-box},
\texttt{radio-button}, \texttt{toggle-button}, \texttt{input} &
Elements where users directly input or modify values; crucial for operation. \\
\hline
High & 10 &
\texttt{push-button}, \texttt{link}, \texttt{menu-item}, \texttt{button} &
Important interactive elements that trigger actions or navigation. \\
\hline
Medium & 20 &
\texttt{heading} &
Indicates content structure; prioritized over simple static text. \\
\hline
Low & 30 &
\texttt{static}, \texttt{image}, \texttt{group}, others &
Elements for information display only; removed when overlapping with interactive elements. \\
\hline
\end{tabularx}
\caption{Tag Priority for UI Element Preservation.
Lower scores indicate higher priority, while higher scores correspond to elements that are more likely to be removed.}\label{tab:tag_priority}
\end{table*}

\nbf{Execution Logic}
For each pair of duplicate candidates, we execute the merge according to the following rules:

\begin{itemize}
    \item \textbf{Priority Comparison:} Based on the scores in Table~\ref{tab:tag_priority}, the element with the higher priority (lower score) is retained, and the lower priority element is removed.
    \begin{quote}
        \textit{Example:} If a \texttt{push-button} "OK" (Score 10) overlaps with a \texttt{static} "OK" (Score 30), the \texttt{push-button} is retained.
    \end{quote}
    \item \textbf{Special Exception:} For a pair consisting of a \texttt{link} and a \texttt{static} element, the \texttt{link} is forcibly prioritized.
    \item \textbf{Tie-breaking:} If both elements have the same priority score (e.g., both are \texttt{static}), the element with the longer label string is retained to preserve more information.
\end{itemize}

\subsection{Paragraph Compression}
\label{app:paragraph_compression}

Long text passages with low relevance to the user's task instruction act as noise that increases the inference load on the agent.
To address this, we introduce a method to dynamically summarize and filter long content, such as UI elements containing the \texttt{Paragraph} tag, based on the instruction content.

\nbf{Preprocessing and Keyword Extraction}
We apply the following normalization steps to both the user instruction and the target text:

\begin{enumerate}
    \item \textbf{Normalization:} Convert all strings to lowercase.
    \item \textbf{Tokenization:} Replace non-alphanumeric characters with whitespace and split the string into a list of words.
    \item \textbf{Filtering:} Remove words found in the general stop-word list ($\mathcal{S}_{\mathrm{stop}}$) and extract only words with a length of 2 or more characters to construct the keyword set $L$.
\end{enumerate}

The stop-word list $\mathcal{S}_{\mathrm{stop}}$ used in this process is defined in Table~\ref{tab:stop_words}. It includes general function words, task-specific conversational fillers, and common generic UI terms.

\begin{table}[h]
    \centering
    \small
    \begin{tabularx}{\linewidth}{@{} l X @{}}
        \toprule
        \textbf{Category} & \textbf{Words} \\
        \midrule
        General Function Words & 
        the, a, an, in, on, at, to, for, of, with, by, from, is, are, am, be, this, that, it \\
        \midrule
        Task Expressions & 
        please, can, could, would, you, i, my, me, need, want, try, make, let \\
        \midrule
        UI Operations \& Generic Nouns & 
        click, tap, press, hit, select, choose, open, go, browse, navigate, find, search, check, uncheck, button, link, tab, menu, window, page, website, site, input, enter, type, fill, text, box, field \\
        \bottomrule
    \end{tabularx}
    \caption{List of Stop Words for Keyword Extraction.}
    \label{tab:stop_words}
\end{table}

\nbf{Context Extraction Based on Keywords}
We search the target text (e.g., \texttt{Paragraph} content) for any words contained in the keyword set $L$.

\begin{itemize}
    \item \textbf{Match Found:}
    When a keyword is found, we identify the index of its first occurrence.
    We then extract a window of a fixed number of characters (default: 50 characters) before and after the keyword, formatting the result as ``\texttt{... [extracted text] ...}''.
    This preserves the context relevant to the instruction while reducing the overall length.
    
    \item \textbf{No Match:}
    If no words from the instruction are found in the text, we retain only the first $N_{\mathrm{max}}$ characters (default: 100 characters) and truncate the rest (e.g., ``\texttt{First 100 chars...}'').
\end{itemize}

This dynamic filtering enables the agent to avoid overlooking critical information relevant to the instruction while preventing context overflow and increased inference costs caused by lengthy texts.

%%%%%%%%%%%%%%%%%%%%%%%%%%%%%%%%%%%%%%%%%%%%%%%

\section{Details of Semantic Structuring and Region Segmentation}
\label{app:semantic_structuring}
This section provides our implementation of the semantic structuring phase of A11y-Compressor, including region segmentation and domain-specific optimizations.

\subsection{Element Reordering.}
UI elements are reordered based on their center coordinates $(c_x, c_y)$, as the order in a linearized a11y tree does not necessarily reflect the visual layout.  
Elements are sorted primarily from top to bottom (Y-axis) and secondarily from left to right (X-axis), producing a sequence aligned with the visual reading order.

\subsection{Region Segmentation.}
We classify UI elements into predefined semantic regions using coordinate information, element tags, and application-specific heuristics.  
Each region represents a high-level functional unit (e.g., \texttt{CONTENT}, \texttt{MODAL}) and is associated with structural and interaction-related properties.

\subsection{Intra-region Structuring.}
Within each region, UI elements are structured using inter-element spatial distances and heading tags.  
A special token \texttt{[BLOCK]} is inserted when the distance between adjacent elements exceeds a threshold $\Theta$, enabling the encoding of two-dimensional layout information into a linear sequence.

\subsection{Domain-specific Optimization.}
For application domains with high information density, we apply additional token-efficiency optimizations.  
For example, in spreadsheet applications (e.g., LibreOffice Calc), we retain only value-containing cells, header cells, and instruction-relevant cells.  
These elements are grouped using row and column indices to reconstruct a structured representation that preserves the tabular layout.

\subsubsection{Google Chrome}
For web browsers, we strictly separate the page content from the browser's native UI.

\noindent\textbf{Region Definitions:}
\begin{itemize}
    \item \texttt{BROWSER\_TABS}: The tab area at the top ($Y < 150\text{px}$) containing specific anchors like "new tab" or "close".
    \item \texttt{ADDRESS\_BAR}: The area containing the URL bar and navigation buttons (Back, Reload, etc.) ($Y < 110\text{px}$).
    \item \texttt{BOOKMARK\_BAR}: The bookmark area located directly below the address bar ($110\text{px} < Y < 150\text{px}$).
    \item \texttt{PAGE\_CONTENT}: The main content area of the rendered web page.
    % \item \texttt{MODAL}: Detected pop-ups or dialogs.
\end{itemize}

\subsubsection{VS Code}
To handle the complex pane structure characteristic of IDEs, we perform detailed region segmentation based on precise coordinate thresholds.

\noindent\textbf{Region Definitions:}
\begin{itemize}
    \item \texttt{APP\_LAUNCHER}: The OS launcher area on the far left ($X \le 5\%$).
    \item \texttt{MENUBAR}: The top menu bar ($Y \le 12\%$).
    \item \texttt{ACTIVITY\_BAR}: The icon bar on the left side ($2\% \le X \le 8\%$).
    \item \texttt{SIDE\_BAR}: The side panel containing the file explorer, etc. ($X \le 30\%$).
    \item \texttt{TAB\_BAR}: The editor tab area ($7\% \le Y \le 16\%$).
    \item \texttt{BREADCRUMB}: The breadcrumb list directly below the tabs ($10\% \le Y \le 18\%$).
    \item \texttt{STATUSBAR}: The status bar at the bottom ($Y \ge 96\%$).
    \item \texttt{CONTENT}: The main editor text area (regions other than the above).
\end{itemize}

\subsubsection{Thunderbird}
As a mail client, Thunderbird requires advanced processing that dynamically switches segmentation logic depending on the active View (e.g., the 3-pane "Mail View" vs. the "Settings View").

\noindent\textbf{Basic Region Definitions:}
\begin{itemize}
    \item \texttt{SPACES\_BAR}: The function switching bar on the far left ($X < 115\text{px}$).
    \item \texttt{FOLDER\_TREE}: The mail folder tree structure ($115\text{px} \le X < 400\text{px}$).
    \item \texttt{MESSAGE\_LIST}: The list of emails (center of screen, $X < 55\%$).
    \item \texttt{PREVIEW}: The email body preview pane (right side, $X \ge 55\%$).
    \item \texttt{TOOLBAR}: The search bar and operation buttons at the top.
\end{itemize}

\noindent\textbf{View-Specific Logic:}
\begin{itemize}
    \item \textbf{Mail View:} The output is structured as a 3-part split (Folder List, Mail List, Preview). The boundary of the mail list (\texttt{SPLIT\_MSG\_LIST\_X}) is dynamically estimated from the element layout.
    \item \textbf{Settings View:} When a settings screen is detected, it is split into a "Settings Category (Sidebar)" on the left and "Settings Items (Main)" on the right. Furthermore, we apply a process to simplify "off-screen items" based on the scroll position.
\end{itemize}

\subsubsection{GIMP}
This configuration handles the multi-window and docking interface typical of image editing software.

\noindent\textbf{Region Definitions:}
\begin{itemize}
    \item \texttt{TOOLBOX}: The toolbox area on the left ($X < 22\%$).
    \item \texttt{DOCKS}: The dock area (layers, brushes, etc.) on the right ($X > 78\%$).
    \item \texttt{CANVAS}: The central image editing area.
    \item \texttt{MENUBAR}: The top menu ($Y < 10\%$).
    \item \texttt{STATUSBAR}: The bottom information bar ($Y > 95\%$).
\end{itemize}

\subsubsection{LibreOffice Suite}
While sharing a common framework, specific regions are defined for Calc, Impress, and Writer respectively, and when region definitions overlap with the common regions, the application-specific definitions take precedence.

\noindent\textbf{Common Semantic Regions:}
\begin{itemize}
    \item \texttt{MENUBAR}: The top menu strip containing "File", "Edit", etc. ($Y < 10\%\sim20\%$).
    \item \texttt{TOOLBAR}: The area immediately below the menubar containing buttons and combo boxes ($Y < 25\%$).
    \item \texttt{STATUSBAR}: The information bar at the bottom of the window ($Y > 90\%\sim96\%$).
\end{itemize}

\noindent\textbf{LibreOffice Calc:}
\begin{itemize}
    \item \texttt{FORMULA\_BAR}: Area below the menu containing the formula bar ($9\% < Y < 23\%$).
    \item \texttt{SHEET}: The spreadsheet cell area.
    \item \texttt{SHEET\_TABS}: Sheet switching tabs at the bottom ($93\% < Y < 96\%$).
\end{itemize}

\noindent\textbf{LibreOffice Impress:}
\begin{itemize}
    \item \texttt{SLIDE\_LIST}: Slide overview on the left ($X < 20\%$).
    \item \texttt{PROPERTIES}: Property panel on the right ($X > 80\%$).
    \item \texttt{CONTENT}: The central slide editing view.
\end{itemize}

\noindent\textbf{LibreOffice Writer:}
\begin{itemize}
    \item \texttt{CONTENT}: The central document editing area.
    \item \texttt{PROPERTIES}: The properties sidebar (if active).
\end{itemize}

\subsubsection{VLC Media Player}
Due to its simple UI structure, we apply a basic vertical split.

\noindent\textbf{Region Definitions:} \\
\begin{itemize}
    \item \texttt{MENUBAR}: Top menu ($Y < 10\%$).
    \item \texttt{TOP\_BAR}: Toolbar area ($Y < 20\%$).
    \item \texttt{CONTENT}: Video playback screen or playlist.
    \item \texttt{STATUSBAR}: Playback controls and seek bar at the bottom ($Y > 92\%$).
\end{itemize}

\subsubsection{OS (Ubuntu)}
Since this handles the entire desktop environment, segmentation is performed without reliance on specific window frames.

\noindent\textbf{Region Definitions:}
\begin{itemize}
    \item \texttt{TOP\_BAR}: System bar at the very top ($Y < 5\%$).
    \item \texttt{APP\_LAUNCHER}: Dock/Launcher on the far left ($X < 6\%$).
    \item \texttt{DESKTOP\_ICONS}: Desktop icon arrangement recognized as a grid.
    \item \texttt{OS\_POPUP}: Right-click menus or dialogs on the desktop.
    \item \textbf{Window Detection:} We dynamically identify window regions on the screen by pairing labels with input forms and analyzing the placement of "Close/Minimize" buttons, treating each as an independent context.
\end{itemize}

% Case study 
\begin{figure*}[!t]
    \centering
    \includegraphics[width=\textwidth]{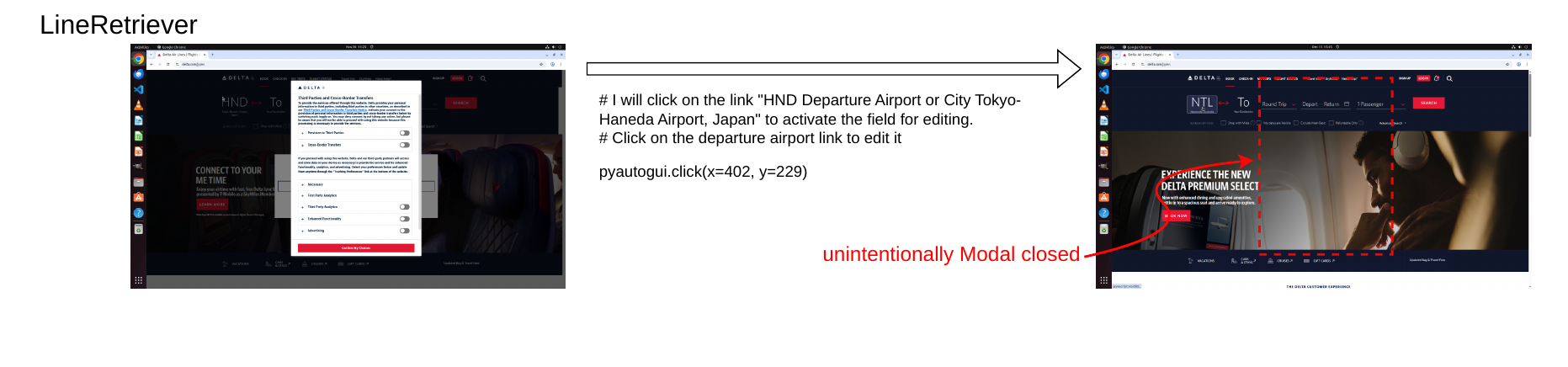}
    \caption{
        Example of modal dialog handling with LineRetriever. In this instance, the model failed to perceive the active modal due to the limited extraction range of the retriever model. When background mouse interactions failed, the model's fallback to keyboard navigation inadvertently reloaded the page, which closed the modal unintentionally.
    }
    \label{fig:case_study}
\end{figure*}

\subsection{Determination of the Threshold $\Theta$}
\label{Theta}
To adaptively handle varying information densities across different applications, we dynamically determine the block segmentation threshold $\Theta$ for each screen.
First, we estimate a base vertical gap, denoted as $G_{base}$, which represents the standard line height or margin of the interface.
Based on this estimate, we then select an optimal multiplier to derive the final threshold $\Theta$ used for block segmentation.

\subsubsection{Estimating the Base Gap}
We begin by computing the vertical distances between all adjacent UI elements in the content area.
Let $dy$ denote the vertical distance between two neighboring elements.
To avoid the influence of large structural gaps, such as margins separating distinct sections, we focus only on the lower 70\% of the resulting distance values.
The base vertical gap, denoted as $G_{base}$, is then defined as the median of this subset.
Finally, $G_{base}$ is clamped to a minimum value to prevent excessive sensitivity to small spacing variations:

\[
G_{base} = \max(\text{median}(dy_{0..70\%}), 40 \text{ pixels}).
\]

\subsubsection{Adaptive Threshold Selection}
The segmentation threshold is defined as $\Theta = G_{base} \times N$. We select the multiplier $N$ adaptively from a candidate set $\{3.0, 4.0, 8.0\}$ to balance structural grouping against fragmentation.

The algorithm iteratively tests these multipliers starting from the strictest value ($3.0$, which yields the lowest threshold). Candidate $\Theta$ is rejected if it causes \textbf{over-segmentation}, defined by the following criteria:
\begin{itemize}
    \item The total number of blocks exceeds 50.
    \item The total number of blocks exceeds 10, and more than 50\% of them contain only a single UI element (indicating that the threshold is too low and fragments otherwise coherent lines).
\end{itemize}

The final $\Theta$ is set to the first multiplier $N$ that satisfies these stability conditions, or to the largest multiplier ($8.0$) if all stricter options fail.
This ensures that $\Theta$ remains robust against varying layout densities, grouping related elements while correctly separating distinct logical blocks.

%%%%%%%%%%%%%%%%%%%%%%%%%%%%%%%%%%%%%%%%%%%%%%%

\section{Case Study: LineRetriever}
\label{app:Case_study}

With LineRetriever~\cite{kerboua2025lineretriever}, the subset of elements extracted by the retriever model caused the agent to be unaware of the modal dialog, leading it to attempt interactions with background elements. However, because the active modal blocked background operations, these attempts resulted in no visual changes on the screen. Consequently, the model switched its strategy from mouse interactions to keyboard operations in an attempt to focus on the target field. This inadvertently shifted the focus to the browser's address bar, triggering a page reload. As a result, the modal was closed unintentionally.

\end{document}